\documentclass[10pt,twocolumn,letterpaper]{article}

\usepackage{cvpr}              

\usepackage[accsupp]{axessibility}  
\usepackage[T1]{fontenc}
\usepackage{graphicx}
\usepackage{pifont}
\usepackage{amsmath}
\usepackage{booktabs}
\usepackage{algorithm}
\usepackage{algorithmic}
\usepackage{enumitem}
\usepackage{amsfonts}
\usepackage{mathabx}
\usepackage{bm}
\usepackage{adjustbox}
\usepackage{url}
\usepackage{listings}
\usepackage{xcolor} 

\definecolor{cvprblue}{rgb}{0.21,0.49,0.74}
\usepackage[pagebackref,breaklinks,colorlinks,citecolor=cvprblue]{hyperref}

\def\paperID{11179} 
\def\confName{CVPR}
\def\confYear{2025}

\title{SACB-Net: Spatial-awareness Convolutions for Medical Image Registration}
\author{ {Xinxing Cheng$^{1}$, Tianyang Zhang$^{1}$, Wenqi Lu$^2$, Qingjie Meng$^{1,5}$,} \\{ Alejandro F. Frangi$^{3,4}$, Jinming Duan$^{1,3,4}$\thanks{Corresponding author.}}\\
$^1$ School of Computer Science, University of Birmingham, UK \\
$^2$ Department of Computing and Mathematics, Manchester Metropolitan University, UK \\
$^3$ Division of Informatics, Imaging and Data Sciences, University of Manchester, UK\\
$^4$ Centre for Computational Imaging and Modelling in Medicine, University of Manchester, UK\\
$^5$ Department of Computing, Impeiral College London, UK\\
{\tt\small \{xxc142, txz009\}@student.bham.ac.uk, \tt\small W.Lu@mmu.ac.uk, m.qingjie@bham.ac.uk}\\
{\tt\small alejandro.frangi@manchester.ac.uk, \tt\small  jinming.duan@manchester.ac.uk}
}

\begin{document}
\maketitle


\begin{abstract}
Deep learning-based image registration methods have shown state-of-the-art performance and rapid inference speeds. Despite these advances, many existing approaches fall short in capturing spatially varying information in non-local regions of feature maps due to the reliance on spatially-shared convolution kernels. This limitation leads to suboptimal estimation of deformation fields. In this paper, we propose a 3D Spatial-Awareness Convolution Block (SACB) to enhance the spatial information within feature representations. Our SACB estimates the spatial clusters within feature maps by leveraging feature similarity and subsequently parameterizes the adaptive convolution kernels across diverse regions. This adaptive mechanism generates the convolution kernels (weights and biases) tailored to spatial variations, thereby enabling the network to effectively capture spatially varying information. Building on SACB, we introduce a pyramid flow estimator (named SACB-Net) that integrates SACBs to facilitate multi-scale flow composition, particularly addressing large deformations. Experimental results on the brain IXI and LPBA datasets as well as Abdomen CT datasets demonstrate the effectiveness of SACB and the superiority of SACB-Net over the state-of-the-art learning-based registration methods.  The code is available at \url{https://github.com/x-xc/SACB_Net}.
\end{abstract}

\section{Introduction}
\label{sec:intro}
\begin{figure}[ht]
    \centering
    \includegraphics[width =0.99\linewidth]{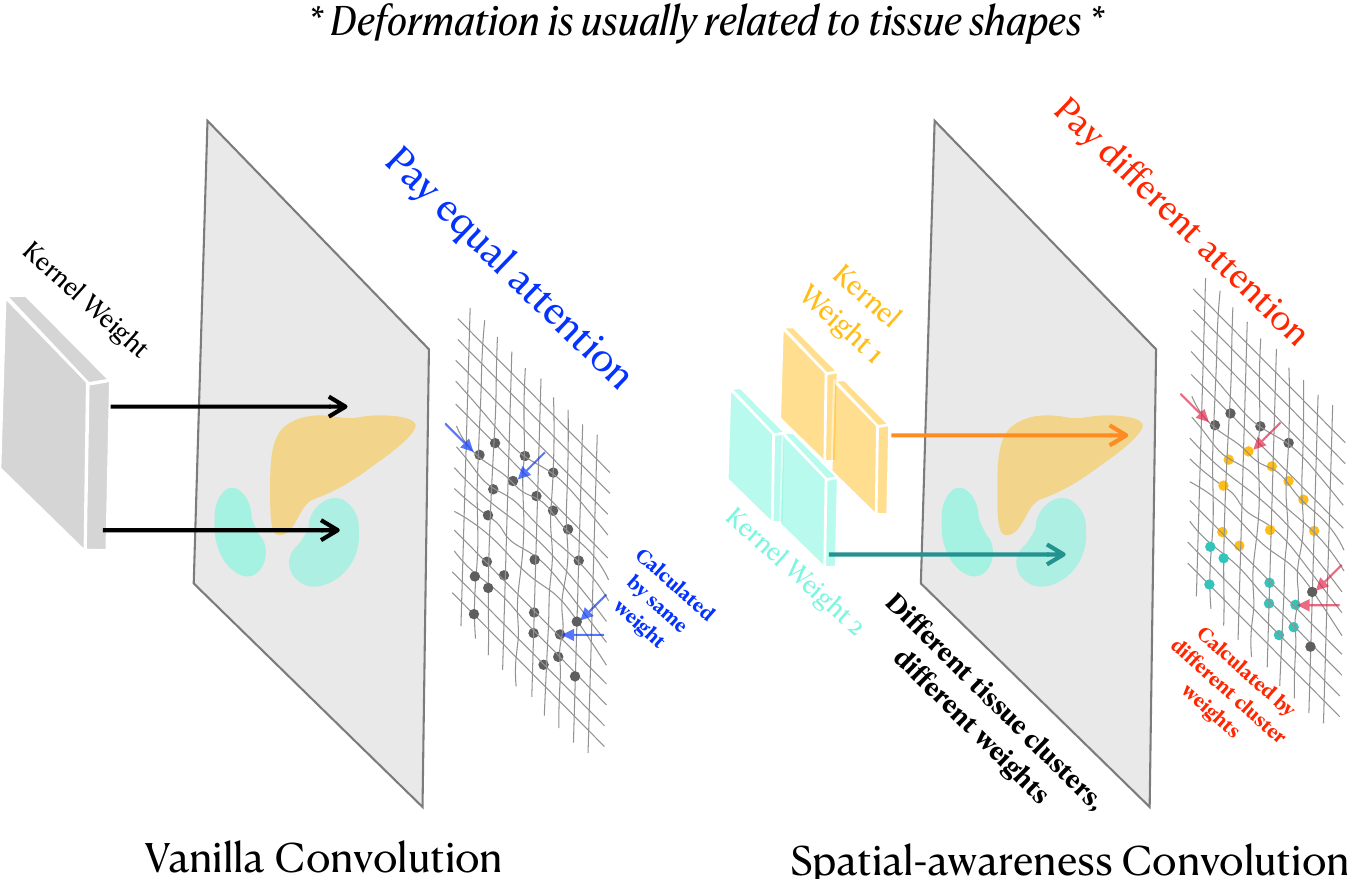}
    \caption{\textbf{Spatial-awareness Convolution.} Since deformation is usually related to tissue shape in medical image registration, voxels/features from different regions should be given varying levels of attention. However, vanilla convolution methods apply shared kernel weights across all regions, leading to a suboptimal estimation of deformation fields. SAC mechanisms, on the other hand, apply distinct kernel attention weights for different spatial clusters.}
    \label{fig:fig1}
\end{figure}
Deformable image registration aims to align the corresponding anatomical structures between a moving and fixed image pair with the estimated dense non-linear deformation field, which plays a crucial role in various medical imaging tasks \cite{sotiras2013deformable,yang2014automated}.
In the past, conventional methods like LDDMM \cite{Beg2005}, DARTEL \cite{DARTEL}, SyN \cite{pmid17659998}, Demons \cite{VERCAUTEREN2009S61}, and ADMM \cite{thorley2021nesterov} are time-consuming and computationally intensive, primarily due to the instance-level (pair-wise) iterative optimization required. Moreover, such methods may involve sophisticated hyper-parameter tuning, limiting their applications in large-scale volumetric registration.
To overcome these challenges, learning-based image registration approaches have demonstrated significant success in achieving competitive accuracy along with fast inference speed. Learning-based approaches can be divided into supervised and unsupervised categories. Supervised methods \cite{cao2018deformable,rohe2017svf,yang2017quicksilver} often rely on the high quality of the synthetic ground truth deformation field or the annotated map \cite{sang2020imposing} to predict a deformation field.
Therefore, unsupervised image registration is more popular for estimating an optimal deformation field of pairwise images by minimizing the distance between the warped moving image and the fixed image with a regularization term. 


Following the classical unsupervised learning-based framework VoxelMorph \cite{balakrishnan2019voxelmorph}, some U-Net-based methods focus on the performance improvements by adding stronger constraints over the deformation field such as inverse/cycle consistency  \cite{zhang2018inverse,kim2021cyclemorph}, gradient inverse consistency \cite{tian2023gradicon} and diffeomorphisms \cite{dalca2018unsupervised,dalca2019unsupervised,mok2020fast}. Some existing methods have introduced more advanced neural architectures, such as large-kernel network \cite{jia2022u}, vision transformers \cite{chen2021vit,chen2022transmorph,ghahremani2024h,shi2022xmorpher,mok2022affine,zhang2021learning}.
To enhance the efficiency of the U-Net architecture and accelerate inference speed, several model-driven methods, including B-Spline \cite{qiu2021learning}, Fourier-Net \cite{jia2023fourier} and WiNet \cite{cheng2024winet}, have been proposed to learn a low-dimensional representation of displacement. The lightweight approach, NCA-Morph \cite{ranem2024nca}, utilizes Neural Cellular Automata to improve the efficiency of image registration.
Another track of recent works tends to progressively estimate the large and complex deformations using either multiple cascades \cite{de2019deep,zhao2019unsupervised,zhao2019recursive,mok2020large} 
or pyramid methods \cite{modelT,hu2022recursive,kang2022dual,ma2024iirp,meng2024correlation,liu2022coordinate} compositions of multi-scale flows derived from multi-scale features. The cascaded methods and pyramid coarse-to-fine methods are capable of effectively improving registration accuracy for estimating large deformation. 


However, existing learning-based methods typically rely on networks that learn local feature correlations by applying shared convolution across all spatial dimensions within each layer. 
As illustrated in Fig. \ref{fig:fig1}, deformations are often linked to the morphological characteristics of tissue in medical image registration, where voxels/features from various anatomical regions need to be paid different attention.  However, plain convolutional methods utilize shared kernel weights uniformly across all regions, resulting in suboptimal estimation of deformation fields. Specifically, 3D medical images encapsulate complex intensity information relevant to anatomy, which makes it hard to capture spatially varying information in non-local regions of feature maps via spatially-shared convolution kernels. This key challenge is effectively incorporating spatially adaptive information, which can be addressed using region-based adaptive convolution techniques. Inspired by \cite{duan2024content}, which explores content-adaptive convolution in 2D image pansharpening, we extend this approach to 3D for medical image registration. This adaptation improves spatial awareness by modifying convolutional kernels based on regional content, thereby enhancing registration accuracy.



In this paper, we propose a novel 3D pyramid medical image registration network, named SACB-Net, which integrates spatial-awareness convolution block (SACB) with a similarity matching module to form a multi-scale flow estimator. The proposed SACB employs the unsupervised KMeans to cluster the learned feature maps. Each feature clustering is believed to capture a specific intensity or structural pattern of the feature maps and then subsequentially used to learn the spatial-aware adaptive convolution kernels. The adaptive kernel generation in SACB allows the network to effectively learn refined feature representations for more accurate flow estimation. We summarize our main contributions as follows:
\begin{itemize}
    \item  We consider the spatial variance between two input images in the context of image registration, which is overlooked by vanilla convolutional-based approaches. The adaptive convolutions in our SACB-Net consider the changing of spatial regions to enhance registration accuracy.  To the best of our knowledge, our method is the first work that employs 3D spatial-awareness convolution in 3D medical registration. 
    \item  The proposed SACB aims to boost the moving and fixed feature maps using adaptive convolution kernels for different spatial regions, where the spatial regions are divided based on the clustering of features, and the convolution kernels are learned from the centroids of each feature clustering. 
    \item We propose an innovative pyramid flow estimator that integrates SACB and similarity matching module to perform multi-scale flow estimation, which can not only estimate local small deformation but also handle large deformation.
    \item Experimental results on atlas-based and inter-subject registration using two brain datasets (IXI and LPBA) demonstrate the effectiveness of SACB-Net. Further evaluation on inter-subject abdominal CT registration shows improved performance in handling large deformations.
\end{itemize}

\section{Related Works}
\subsection{U-Net-based Registration Method} 
VoxelMorph \cite{balakrishnan2019voxelmorph, balakrishnan2018unsupervised}, a general framework for learning-based image registration, demonstrates the effectiveness of unsupervised learning in pairwise image registration using a simplified U-Net architecture. Building on this foundation, recent methods have enhanced registration performance by incorporating stronger constraints on the deformation field, such as inverse/cycle consistency \cite{zhang2018inverse, kim2021cyclemorph}, gradient inverse consistency \cite{tian2023gradicon}, and diffeomorphism constraints \cite{dalca2018unsupervised, dalca2019unsupervised, mok2020fast}. Another track of methods tried to capture long-range dependencies of feature correlations using either large kernel convolution \cite{jia2022u} or vision transformer \cite{chen2021vit,chen2022transmorph,ghahremani2024h,shi2022xmorpher,mok2022affine,zhang2021learning}. TransMorph \cite{chen2022transmorph} incorporates Swin transformer blocks in the encoder, improving the network’s capacity to capture long-range dependencies, though this comes with increased computational cost. LKU \cite{jia2022u}, on the other hand, employs a large kernel convolutional encoder to model long-range relationships with fewer parameters and reduced computational load. NCA-Morph \cite{ranem2024nca} employs Neural Cellular Automata mechanisms, drawing inspiration from biological processes, to improve registration efficiency. While U-Net-based methods have shown comparable performance for small deformations, they struggle to accurately estimate large deformations or register images that are not well pre-aligned.

\subsection{Model-driven Registration Method}
To reduce redundant convolution operations in the U-Net decoder and enhance training and inference efficiency, model-driven approaches such as B-Spline \cite{qiu2021learning} and Fourier-Net \cite{jia2023fourier} focus on learning a low-dimensional representation of the displacement field. However, displacement fields are not inherently smooth, and as a result, methods that rely solely on low-frequency components \cite{jia2023fourier} or interpolate displacement from regular control points \cite{qiu2021learning} may fail to fully and accurately capture the entire displacement field. To address this limitation, WiNet \cite{cheng2024winet} introduces a wavelet-based decoder to preserve high-frequency information in the deformation field while maintaining network efficiency. Nonetheless, similar to U-Net-based methods, model-driven approaches also face challenges in effectively estimating large and complex deformations.
\subsection{Cascaded and Pyramid-based Method} Recent works that estimate large deformation can be either network cascade  \cite{de2019deep,zhao2019unsupervised,zhao2019recursive,mok2020large} or coarse-to-fine pyramid \cite{modelT,hu2022recursive,kang2022dual,ma2024iirp,meng2024correlation,liu2022coordinate}. LapIRN \cite{mok2020large} utilizes a Laplacian pyramid network to capture large deformations by cascading registered flows at three different scales. 
PRNet++ \cite{kang2022dual} utilizes a dual-stream pyramid architecture for coarse-to-fine flow estimation, performing sequential warping on multi-scale feature maps and incorporating local 3D correlation layers. Im2grid \cite{liu2022coordinate} embedded the coordinate information on extracted multi-scale feature maps and estimated the scale flow by calculating the matching score on the neighborhood feature. ModeT \cite{modelT} leverages neighbourhood attention mechanisms to estimate multi-head, multi-scale sub-deformation flows from pyramid feature maps. It then applies weighting modules to fuse these multi-head flows at each scale, generating the final deformation by sequential composition. To solve the insufficient decomposition problems, RDN \cite{hu2022recursive} compose the multi-scale flows by performing level-wise recursion at each resolution level and stage-wise recursion at feature pyramid level.
\subsection{Adaptive Convolution}
Unlike standard convolution operators that use shared convolution kernels, adaptive convolution operators apply different kernels based on variations in the input, providing enhanced feature extraction capabilities and greater flexibility. Following \cite{jia2016dynamic}, recent methods have applied adaptive convolution, demonstrating its effectiveness in various 2D computer vision tasks, such as image upsampling  \cite{su2019pixel}, classification \cite{chen2020dynamic,Zhou_2021_CVPR}, segmentation \cite{chen2021dynamic} and remote sensing pansharpening \cite{jin2022lagconv,duan2024content}. However, spatially adaptive convolution has not yet been explored in 3D medical image registration tasks.

\begin{figure*}[!ht]
    \centering    \includegraphics[width=1\linewidth]{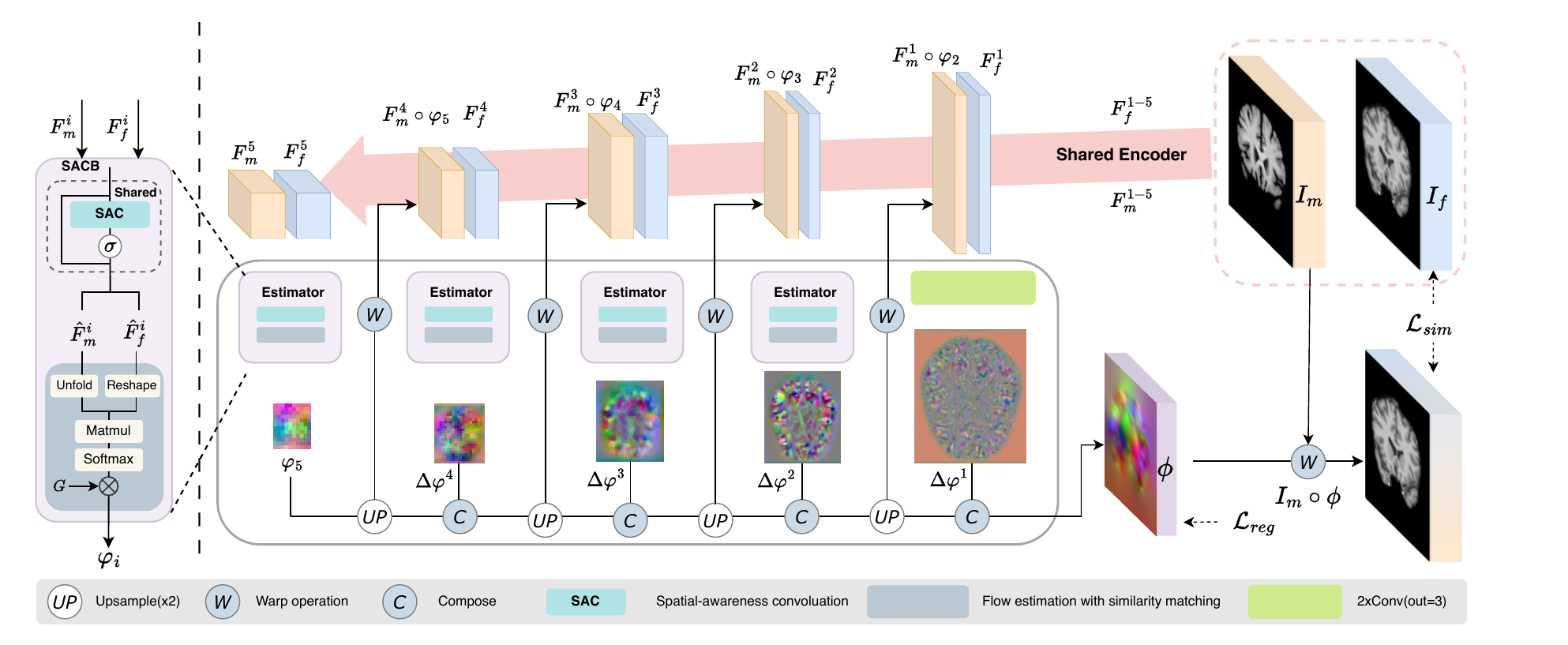}
    \caption{Illustration of a 5-level pyramid SACB-Net. SACB-Net includes a shared encoder that extracts multi-scale feature maps $\{F_m^i\}$ and $\{F_f^i\}$  for the moving image $I_m$ and the fixed image $I_f$, as well as pyramid flow estimators at each scale. At the lowest level, the flow estimator learns deformation ($\varphi_5$) from the extracted moving and fixed image features ($F_m^5$ and $F_f^5$). The following flow estimators take the level-wise features and the output deformation from its preceding level to compose the deformation. Each flow estimator includes a Spatial-Awareness Convolution Block (SACB) to enhance spatially adaptive feature representation, along with a similarity matching module for flow estimation. }
    \label{fig:f2}
\end{figure*}
\section{Method}
\subsection{Network Overview}
Given a pair of moving image $I_m$ and fixed image $I_f$ over the spatial domain $\Omega \subset \mathcal{R}^3$, the SACB-Net$(I_m, I_f; \boldsymbol{\theta})$ network aims to learn the optimal parameters $\boldsymbol{\theta}_{\ast}$ that estimate a deformation field $\boldsymbol{\phi}_{m\rightarrow f}: \mathcal{R}^3\rightarrow\mathcal{R}^3$, mapping the corresponding coordinates between the paired images. In unsupervised setting, the loss function contains $\mathcal{L}_{{sim}}(\cdot)$  measuring the similarity between the warped moving image $I_m \circ \boldsymbol{\phi}$ and $I_f$, and the regularization term. In our experiments, the smoothness term  $\mathcal{L}_{{reg}}(\boldsymbol{\phi}) = \left\|\nabla \boldsymbol{\phi}(\boldsymbol \theta)\right\|_2^2 $ is computed using the $L_2$-norm of the gradient of the deformation field.
The objective function of our network is 
\begin{align}
    \mathcal{L}(\boldsymbol{\theta})= \mathcal{L}_{sim}(I_m \circ(\boldsymbol{\phi}(\boldsymbol \theta)+\mathrm{Id}), I_f) +\lambda \mathcal{L}_{{reg}}
\end{align}
 where $\boldsymbol{\theta}$ denotes the learnable network parameters, $\circ$ represents the warping operator, $\mathrm{Id}$ is the identity grid, $\nabla$ refers to the first-order gradient, and $\lambda$ is the weight of the regularization term. 
We employ Normalized Cross-Correlation (NCC) to measure the similarity.


As illustrated in Fig. \ref{fig:f2}, SACB-Net consists of a shared encoder and a pyramid flow estimator. The shared encoder is designed to extract multi-scale features \( \{F_m^{i} \}\) and \( \{F_f^{i}\} \) for the moving image \( I_m \) and the fixed image \( I_f \), respectively, in a pyramid structure. Here, \( i \in \{1,2,3,4,5\} \) represents the scale of the features, with   resolutions of \( \left[ \frac{D}{2^{i-1}}, \frac{H}{2^{i-1}}, \frac{W}{2^{i-1}} \right] \) for the depth, height and width. The shared encoder consists of five 3D convolutional layers and four down-sampling layers. The convoluational layer is composed of 3D convolution operation, instance normalization \cite{ulyanov2016instance} and LeakyReLU activations \cite{maas2013rectifier} with the $0.1$ slope. Each convolutional layer is followed by an average pooling operation with a kernel size of 2. The extracted features \( \{F_m^{i} \}\) and \( \{F_f^{i}\} \) are then input into our pyramid flow estimators,  where they are used to perform flow estimation and composition as outlined in Sections \ref{sec1} and \ref{sec2}. 

\subsection{3D Spatial-awareness Convolution Block}\label{sec1}
To effectively extract region-based context information for flow estimation, the proposed 3D SAC incorporates prior knowledge of spatial context through a spatial context estimation module and adaptive kernel generator.
 First, the spatial context estimation module identifies spatial regions of interest, producing spatial contextual information. This contextual data is subsequently utilized by the adaptive kernel generator to dynamically produce convolutional kernels tailored to capture non-local, region-specific information. 
\begin{figure*}[!ht]
    \centering
\includegraphics[width=1\linewidth]{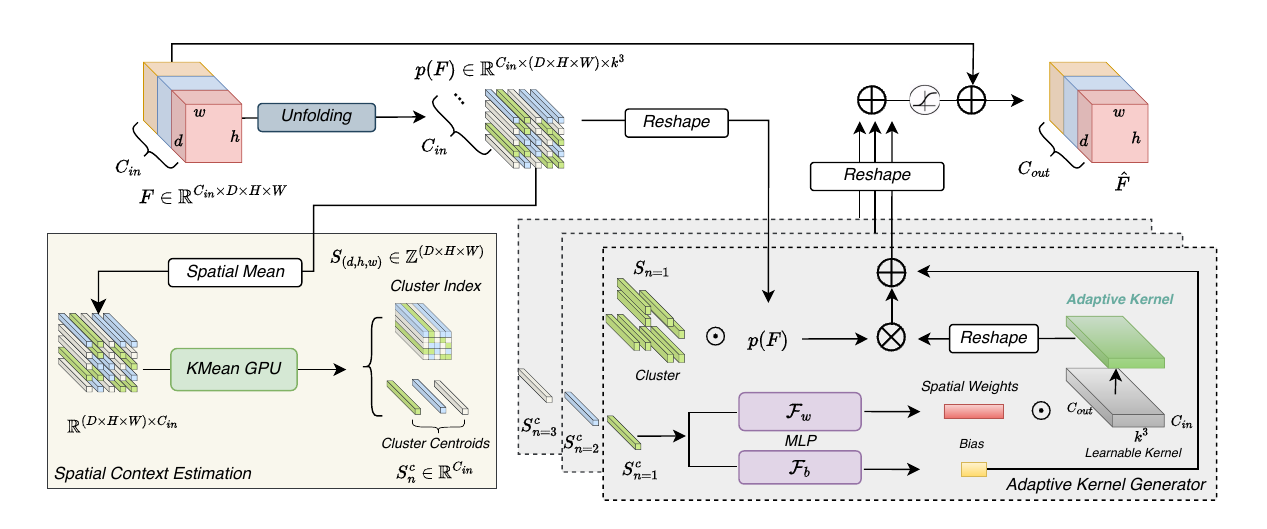}
    \caption{Architecture of the 3D spatial-awareness convolution block (example in three clusters). This block aims to refine the input feature $\mathbf{F}$ to $\hat{\mathbf{F}}$ by adaptive convolution learned from spatial feature clustering. SACB consists of three parts: 1) the spatial context estimation module employs KMeans to cluster similar spatial features on the patched unfolding features, in which the features belonging to the same cluster centroid $S_n^c$ will be indicated with a cluster index map $S_n$. 2) the adaptive kernel generator leverages each cluster centroid to generate cluster-specific spatial weights and bias via two MLPs, the resulting weights and bias will be used to form the final spatial adaptive convolution kernel, which will be imposed on the cluster-indexed unfolding features, again highlighting the spatial-awareness.  3) the original features and spatial-aware features are residually connected to form the final refined $\hat{\mathbf{F}}$. }
    \label{fig:f3}
\end{figure*}

{\smallskip \noindent\bf{Spatial Context Estimation.}} After obtaining the multi-scale features \( \{F_m^{i} \}\) and \( \{F_f^{i}\} \), prior spatial region information is incorporated to refine these features.
To achieve this, we employ unsupervised clustering method K-Means to estimate spatial region for feature maps. The reason not to use label information is due to potential inconsistencies in the labeling of $I_m$  and $I_f$ within the feature space, as well as to address the challenge of limited label availability. Let $\mathbf{F} \in \mathbb{R}^{C_{in} \times D \times H \times W}$ denote the input feature map, where $C_{in}$ represents the number of input channels, and $D$, $H$, and $W$ correspond to the depth, height, and width of the spatial dimensions, respectively. As shown in Fig. \ref{fig:f3}, first, the feature map \( \mathbf{F} \) is unfolded into local patches to capture neighborhood information using sliding windows of size \( k\). A spatial mean is then applied to each patch to reduce its spatial dimensions. For each voxel \( V_{(d,h,w)} \) located at coordinates \( (d,h,w) \) within \( \mathbf{F} \), suppose $p(\cdot)$ is the unfolding operation, the corresponding unfolded patch is defined as:
\begin{align}
p_{(V_{(d,h,w)})}= \left\{ V_{(d + d', h + h', w + w')} \mid d', h', w' \in \left[-\frac{k}{2}, \frac{k}{2}\right] \right\}.
\end{align}
Here, $p_{(V_{(d,h,w)})}$ represents a patch of voxels centered at \( V_{(d,h,w)} \), with offsets \( d', h', w' \) that range from \( -\frac{k}{2} \) to \( \frac{k}{2} \) (assuming \( k \) is odd). The unfolded feature denotes as $p(F) \in \mathbb{R}^{C_{in} \times (D\times H \times W) \times k^3}$. The spatial mean of each local patch  $\overline{p}_{(V_{(d,h,w)})}$ \footnote{The choices of $\overline{p}$ are investigated in Sec.~\ref{sec:ablation}.} based on the neighborhood voxel is calculated as:
\begin{align}\label{p}
    \overline{p}_{(V_{(d,h,w)})} = \frac{1}{k^3} \sum_{d' = -\lfloor \frac{k}{2} \rfloor}^{\lfloor \frac{k}{2} \rfloor} \sum_{h' = -\lfloor \frac{k}{2} \rfloor}^{\lfloor \frac{k}{2} \rfloor} \sum_{w' = -\lfloor \frac{k}{2} \rfloor}^{\lfloor \frac{k}{2} \rfloor}p_{(V_{(d,h,w)})}.
\end{align}
Afterward, the precessed feature \(\mathbf{F}\) is reshaped to dimensions \( [D \times H \times W, C_{in}] \) to estimate a cluster index matrix \( S  \in \mathbb{Z}^{D\times H \times W} \), which assigns each voxel to a cluster \( S_n = \{(d,w,h)| S_{(d,w,h)} = n\}; n \in \{1,\cdots,N\}\). Here, $N$ represents the maximum number of clusters used for GPU-based KMeans clustering. Additionally, the centroid voxel of each cluster  $S_n^c \in \mathbb{R}^{C_{in}}$ is computed as
\begin{align}
    S_n^c = \frac{1}{|S_n|} \sum_{(d,w,h) \in S_n} {V}_{(d,w,h)},
\end{align}
where  $|S_n|$ denotes the total number of voxels belonging to the \( n^{\text{th}} \) spatial region. The cluster index matrix \( S\) and the centroid voxels \( S^c \) for each cluster are then utilized to generate adaptive kernel weights for convolution.


{\smallskip \noindent \bf{Adaptive Kernel Generator.}} To adaptively generate a spatial convolution kernel that incorporates region-based information, we employ a multi-layer perception to model the mapping \( \mathcal{F}_w \), which produces an adaptive convolution kernel \( \mathbf{W}_n \in \mathbb{R}^{C_{out} \times C_{in} \times k^3} \) for the \( n^{\text{th}} \) spatial region based on \( S_n^c \). To simplify the process, we first initialize a global learnable convolution kernel \( \mathbf{W} \in \mathbb{R}^{C_{out} \times C_{in} \times k^3} \) and then use the mapping function \( \mathcal{F}_w \) to generate a spatial weight that can be applied to the global kernel \( \mathbf{W} \). Thus, \( \mathbf{W}_n \) is obtained as
\begin{align}
     \mathbf{W}_n = \mathcal{F}_w(S_n^c) \odot \mathbf{W},
\end{align}
where $\odot$ denotes element-wise dot product.
Then, a mapping function \( \mathcal{F}_b \) using a multi-layer perception is applied to compute the convolution bias  via obtained centroid $S_n^c$. Consequently, the region-based spatially adaptive convolution is represented as:
\begin{align}
   SAC( \mathbf{F}_{(d,w,h)}) =p_{(V_{(d,h,w)})}\otimes \mathbf{W}_n  + \mathcal{F}_b(S_n^c), 
\end{align}
where \( p(\cdot) \)  denotes the unfolding operation outlined in the spatial context estimation section. Subsequently, the process of applying the SAC block to produce refined features is as follows
\begin{align}
    \hat{\mathbf{F}} = SACB(\mathbf{F}) = \mathbf{F} +  \sigma( SAC( \mathbf{F}) ),
\end{align}
where  $ \sigma(\cdot)$ denotes activation function.
\subsection{Pyramid Flow Estimator}\label{sec2}
 Let ${\hat{F_m}^{i}}$ and ${\hat{F_f}^{i}}$ represent the multi-scale features refined by the proposed SACB at the $i^{th}$ level, the sub-deformation flows at each scale are then estimated based on the matching score using the refined features. Specifically, we calculate the similarity matching score between each voxel in the fixed image features and corresponding neighbouring voxels in the moving image features. Similar to \cite{sun2021loftr,liu2022coordinate}, this operation can be efficiently implemented as a matrix operation using the dot product.
For instance, at the coarsest level $i=5$, the matching score can be computed as 
\begin{align}
    M_{{sim}}(\hat{F_f}^{5}, \hat{F_m}^{5}) = \text{Softmax}(\hat{F_f}^{5}p(\hat{F_m}^{5})^T),
\end{align}
where \( p(\cdot) \) represents the unfolding operation as described in Section \ref{sec1}. For each voxel $V$ at $\hat{F_m}^{5}$, the similarity matching score can be calculated as
\begin{align}
    &M_{{sim}}(\hat{F_f}^{5}(V), \hat{F_m}^{5}(V)) = \\ \nonumber
     &\text{Softmax}\left( \left\lbrace \left\langle \hat{F_f}^{5}(V), \hat{F_m}^{5}(V_c) \right\rangle \right\rbrace_{V_c \in  \mathcal{N}(V)} \right),
\end{align}
where \( \langle \cdot, \cdot \rangle \) denotes the inner product, and \( \mathcal{N}(V) \) represents the set of voxels within an \( k \times k \times k \) neighborhood centered around voxel \( V \), $k$ represents the window size of the search region of $\hat{F_m^{5}}$. We set $k=3$ in our experiments. Then, the sub-deformation flow at the \( 5^{\text{th}} \) scale can be obtained as
\begin{align}
    \varphi^5 = M_{{sim}}(\hat{F_f}^{5}, \hat{F_m}^{5})G
\end{align}
$G \in \mathbb{R}^{(k \times k \times k) \times 3}$ is a 3D grid that encodes the relative positions with respect to the centroid voxel location.

At the $i=4$ level, we first up scaling the flow $\varphi^5$ to form  $\hat{\varphi}^5$, then $\hat{F_m}^{4}$ is warpped with $\hat{\varphi}^5$. Sequentially, the $M_{{sim}}$ need to calculated between $\hat{F_f}^{4}$ and $\hat{F_m}^{4}\circ \hat{\varphi}^5$ for obtaining $ \Delta \varphi^4$. Afterward, the overall process can be formulated as:
\begin{equation}
\begin{cases}
\hat{\varphi}^{i} = \text{up}_{2\times}(\varphi^{i}), \\
\hat{F_f}^{i-1} = SACB(F_f^{i-1}),\\
\hat{F_m}^{i-1} =  SACB(F_m^{i-1} \circ \hat{\varphi}^{i} ),\\
\Delta \varphi^{i-1} =  M_{{sim}}(\hat{F_f}^{i-1}, \hat{F_m}^{i-1})G, \\
\varphi^{i-1} = \hat{\varphi}^{i} \circ \Delta \varphi^{i-1}, 
\end{cases}  i \in \{5,4,3,2\}.
\end{equation}
The $\text{up}_{2\times}(\cdot)$ is flow rescaling operation with 2 $\times$ resolution. 

We note that the matching score calculation was not applied for flow estimation at the $1^{st}$ scale, as the search region with $k=3$ can only determine a maximum displacement of 1 voxel using full-resolution features. 
However, the calculations at the  $1^{st}$ scale significantly increase memory and computational costs, yielding limited benefit.
Therefore, we used a simple approach with two convolutional layers for flow estimation at the final scale.

\section{Experiments}
\begin{table*}[!ht]
\centering
    \caption{Registration performance comparison on the IXI and LPBA datasets.}
    \begin{adjustbox}{width=1.0\textwidth}
        \begin{tabular}{lccccccccc}
        \toprule[1pt]
        & & \multicolumn{2}{c}{\textbf{IXI} (30 ROIs)} & & &\multicolumn{2}{c}{\textbf{LPBA}  (54 ROIs)} \\
      \cmidrule(lr){2-5} \cmidrule(lr){6-9}
        Method & Dice$\uparrow$  &HD95$\downarrow$ &ASSD$\downarrow$ &  $|J|_{ < 0}\%$$\downarrow$  & Dice$\uparrow$  &HD95$\downarrow$& ASSD$\downarrow$& $|J|_{ < 0}\%$ $\downarrow$& \# Param \\
        \midrule
        Affine   & 0.386$\pm$0.195 &6.479$\pm$0.666 &2.445$\pm$0.280 &-  &0.525$\pm$0.047 & 8.039$\pm$0.861& 2.586$\pm$0.350&-&- \\
    SyN~\cite{avants_ANTS} & 0.645$\pm$0.152 &6.394$\pm$1.048 &1.551$\pm$0.286 & \textless{}0.0001  &0.707$\pm$0.016 & 6.254$\pm$0.444 &1.479$\pm$0.131&\textless{}0.0001  &-  \\
     \midrule
    VM-1~\cite{balakrishnan2019voxelmorph}  & 0.729$\pm$0.129 &3.798$\pm$0.757 &  0.937$\pm$0.162 & 1.590$\pm$0.339   &0.664$\pm$0.025&6.873$\pm$0.654&1.717$\pm$0.200&0.649$\pm$0.261&\: 0.27M\\
    VM-2~\cite{balakrishnan2019voxelmorph}  &0.732$\pm$0.123&3.723$\pm$0.680 & 0.926$\pm$0.158 &1.522$\pm$0.336  &0.669$\pm$0.025&6.847$\pm$0.659&1.698$\pm$0.200& 0.591$\pm$0.242&\: 0.30M\\
    NCA-Morph~\cite{ranem2024nca} & 0.753$\pm$0.136 & \: 3.109$\pm$0.525& 0.796$\pm$0.121&0.506$\pm$0.190 & 0.679$\pm$0.023&6.666$\pm$0.634&1.631$\pm$0.188&0.264$\pm$0.121 &\: 0.37M \\
    TransMorph~\cite{chen2022transmorph} & 0.754$\pm$0.124  &3.543$\pm$0.721 &0.862$\pm$0.168 & 1.579$\pm$0.328  & 0.695$\pm$0.022&6.564$\pm$0.619 & 1.559$\pm$0.182 & 0.474$\pm$0.176&46.77M\\
    LKU~\cite{jia2022u} &0.765$\pm$0.129 & 2.967$\pm$0.494 &0.757$\pm$0.114  & 0.109$\pm$0.054 &0.706$\pm$0.032&6.452$\pm$0.951&1.603$\pm$0.250&0.594$\pm$0.203&\: 2.09M\\
     \midrule
    B-Spline-Diff~\cite{qiu2021learning} & 0.742$\pm$0.128 & 3.256$\pm$0.538&0.832$\pm$0.117 & \textless{}0.0001  & 0.665$\pm$0.023&6.792$\pm$0.622&1.713$\pm$0.186&\textbf{0.0$\pm$0.0}&\: 0.27M\\
    Fourier-Net~\cite{jia2023fourier} & 0.763$\pm$0.129 &\textbf{2.857$\pm$0.456} &\textbf{0.748$\pm$0.114} & 0.024$\pm$0.019  &0.672$\pm$0.022 & 6.716$\pm$0.601&1.666$\pm$0.180 &0.216$\pm$0.104 &\: 4.20M \\
     \midrule
    LapIRN~\cite{mok2020large}& 0.763$\pm$0.133& 3.166$\pm$0.608 &0.779$\pm$0.126 &0.312$\pm$0.106 &0.716$\pm$0.016&6.116$\pm$0.454&1.426$\pm$0.133&0.024$\pm$0.009& \: 1.20M \\
    PRNet++~\cite{kang2022dual} &0.755$\pm$0.130 & 3.593$\pm$0.748 & 0.857$\pm$0.162 &1.052$\pm$0.302&0.701$\pm$0.021&6.492$\pm$0.597&1.520$\pm$0.177&0.072$\pm$0.027& \: 1.24M\\
    ModeT~\cite{modelT}&0.758$\pm$0.125&3.496$\pm$0.732 &0.828$\pm$0.151&0.114$\pm$0.057 &  0.721$\pm$0.013& 5.969$\pm$0.416&1.375$\pm$0.110&0.010$\pm$0.004&\: 1.03M\\
    Im2Grid~\cite{liu2022coordinate}  &0.761$\pm$0.127 &3.316$\pm$0.668&0.799$\pm$0.128 &\textless{}0.0002 &0.713$\pm$0.014&6.062$\pm$0.428&1.419$\pm$0.118&0.007$\pm$0.003&\: 0.89M\\
    RDN~\cite{hu2022recursive}&0.759$\pm$0.123&3.476$\pm$0.802 &0.823$\pm$0.161  &\textless{}0.0001&0.713$\pm$0.017&6.208$\pm$0.497&1.436$\pm$0.142&\textless{}0.0002&28.65M \\
     \midrule 
    Ours~&\textbf{0.769$\pm$0.127}&3.128$\pm$0.631 &0.760$\pm$0.125&0.083$\pm$0.045& \textbf{0.731$\pm$0.012}&\textbf{5.862$\pm$0.436}& \textbf{1.326$\pm$0.114}& 0.018$\pm$0.006&\: 1.11M \\
            \bottomrule[1pt]
        \end{tabular}

\end{adjustbox}
    \label{tab:tab1}
\end{table*}

\noindent \textbf{Datasets.} 
Our experiments use two publicly available 3D brain MRI datasets IXI\footnote{\url{https://brain-development.org/ixi-dataset/}} and LPBA \cite{shattuck2008construction}, along with one public abdominal CT dataset\footnote{\url{https://learn2reg.grand-challenge.org/}} \cite{xu2016evaluation}. The preprocessed IXI dataset \cite{chen2022transmorph} includes 576 MRI scans (160×192×224) from healthy subjects. We followed the same protocol as \cite{chen2022transmorph}, partitioning the data into 403, 58, and 115 scans for training, validation, and testing sets, respectively. Atlas-based brain registration was conducted using a generated atlas from \cite{kim2021cyclemorph}. The LPBA dataset consists of 40 T1-weighted MRI volumes. Each MRI scan has been rigidly pre-aligned to the MNI305 template and manually annotated with 54 regions of  interest (ROIs). We center-crop all volumes to dimensions of $160\times192\times160$ with voxel size $1\times1\times1mm^3$. LPBA dataset was used to perform inter-subject registration with divided three subsets: 25 volumes (25$\times$24 pairs) for training, 5 (5$\times$4 pairs) for validation, and 10 (10$\times$9 pairs) for testing. The Abdomen CT dataset consists of 30 abdominal scans with 13 manually labelled anatomical structures. We clamp the Hounsfield Units (HU) to  $\left[-1000,1000\right]$ and normalize each scan to $\left[0,1\right]$. Each scan is center-cropped from $192\times160\times256$ to $192\times160\times224$ with voxel size $2\times2\times2$ mm$^3$. The dataset is divided into 19 training volumes (19$\times$18 pairs), 3 validation volumes (3$\times$2 pairs) and 8 testing volumes (8$\times$7 pairs) for inter-subject registration.

{\smallskip \noindent\bf{Implementation Details.}} Our method is implemented using PyTorch. Both the training and testing phases are deployed on an A100 GPU with 40GB VRAM. All models are optimized using Adam followed with the same learning rate $10^{-4}$ and batch size 1. A fast GPU K-Means implementation \footnote{\url{https://github.com/jokofa/torch_KMeans}} is adopted.

{\smallskip \noindent\bf{Evaluation Metrics.}} To qualify the registration performance, we use the Dice similarity coefficient (Dice) to evaluate the region of interest (ROI) segmentation overlap, the average symmetric surface distance (ASSD) \cite{taha2015metrics} and the 95 percentile of the Hausdorff distance (HD95) to measure the registration accuracy of ROI structures. For evaluating the diffeomorphism of deformation fields, we use the percentage of negative values of the Jacobian determinant ($|J|_{ < 0}\%$)
to calculate the percentage of folding voxels.

{\smallskip \noindent\bf{Comparative Methods.}} We compare our method against state-of-the-art approaches, including the conventional SyN method \cite{pmid17659998}, U-Net-based models such as VoxelMorph (VM) \cite{balakrishnan2019voxelmorph}, LKU \cite{jia2022u}, and TransMorph \cite{chen2022transmorph}, model-driven methods like B-Spline-Diff \cite{qiu2021learning} and Fourier-Net \cite{jia2023fourier}, the lightweight NCA-Morph \cite{ranem2024nca}, the cascaded method LapIRN \cite{mok2020large}, and four pyramid-based methods: PR++ \cite{kang2022dual}, Im2Grid \cite{liu2022coordinate}, RDN \cite{hu2022recursive} and ModeT \cite{modelT}. All models were trained using their officially released code, with optimal hyperparameters tuned on the  validation sets.

\section{Results}
In this section, we evaluate our SACB-Net against comparative methods on three tasks: atlas-based brain registration using the IXI dataset, inter-subject brain registration using the LPBA dataset, and large deformation estimation for inter-subject abdominal CT registration.

\subsection{Atlas-based Brain Registration}
Table \ref{tab:tab1} presents the quantitative results of atlas-based brain registration for 30 anatomical structures evaluated on the IXI dataset. It can be seen that SACB-Net outperforms all comparable methods in terms of Dice score while achieving plausible deformations and third-best results on HD95 and ASSD.
Compared to the best-performing U-Net-based method (LKU), our SACB-Net achieves a 0.4\% improvement in Dice score while using 0.98M fewer parameters. In terms of model-driven methods, SACB-Net surpasses Fourier-Net by a margin of 0.6\% in the Dice score and uses 3.09M fewer parameters. For cascade and pyramid-based methods, those methods have a similar number of parameters apart from RDN. Our SACB-Net outperforms the second-best cascade method LapIRN with 0.6\% Dice, 0.038mm HD95, and 0.019mm ASSD, respectively. 

\subsection{Inter-subject Brain Registration}
The inter-subject registration of the LPBA dataset tends to have a slightly larger deformation than the atlas-based registration on the IXI dataset. As shown in Table \ref{tab:tab1}, most UNet-based and model-driven methods do not perform as well as cascade and pyramid-based methods. Only LKU surpasses the pyramid-based method PR++, achieving improvements of 0.5\% in Dice, 0.04 mm in HD95 and 0.083 mm in ASSD, respectively. Our SACB outperforms all comparable methods in Dice, HD95 and ASSD with plausible deformations. Compared to the second-best method, ModeT, SACB achieves improvements of 1\% in Dice, 0.107mm in HD95, and 0.049mm in ASSD, respectively. 
The visualization in Fig. \ref{fig:vis_res1}
shows that our SACB-Net captures more detailed correspondences in the registration process, as highlighted in the red box.
\subsection{Inter-subject Abdomen Registration}
The primary challenge in abdominal CT registration tasks is caused by the substantial differences in the spatial distribution of organs across scans, making it difficult to accurately estimate large deformations during the registration process. Table \ref{tab:ct} demonstrates the numerical results of the inter-subject abdomen registration task. Our results show that our method outperforms all compared approaches in terms of Dice, HD95, and ASSD. Notably, U-Net-based and model-driven methods perform inferior to pyramid and cascaded approaches for estimating large deformations. Even the best-performing method in such approaches, TransMorph,  still exhibits a significant gap compared to our SACB-Net with $14.4\%$ difference in Dice score, 5.934mm in HD95 and 3.153mm in ASSD, respectively. Compared to the second-best pyramid method, ModeT, SACB-Net achieves improvements of 3.8\% in Dice, 2.098 mm in HD95, and 0.781 mm in ASSD, respectively.
\begin{table}[!ht]
\centering
    \caption{Registration performance comparison on Abdomen CT.}
         \begin{adjustbox}{width=\linewidth}
        \begin{tabular}{lccccc}
        \toprule[1pt]
        & &\multicolumn{2}{c}{\textbf{Abdomen CT} (13 ROIs)} \\
        \cmidrule(lr){2-5}
        Method & Dice$\uparrow$  &HD95$\downarrow$& ASSD$\downarrow$& $|J|_{ < 0}\%$ $\downarrow$ \\
        \midrule
        Affine &0.305$\pm$0.047 &27.790$\pm$4.868& 10.869$\pm$2.075 &-\\
     \midrule
    VM-2~\cite{balakrishnan2019voxelmorph} & 0.406$\pm$0.058  &  25.226$\pm$4.942& 8.890$\pm$2.066& 3.882$\pm$1.512\\
    NCA-Morph~\cite{ranem2024nca}&0.406$\pm$0.061&25.959$\pm$4.982&  9.083$\pm$2.223&5.197$\pm$1.166 \\
    TransMorph \cite{chen2022transmorph} & 0.444$\pm$0.065 &24.187$\pm$4.955&8.156$\pm$2.065& 6.045$\pm$1.750\\
    LKU~\cite{jia2022u} &0.423$\pm$0.058 &24.252$\pm$4.604&8.395$\pm$1.946&4.316$\pm$1.778 \\
    \midrule
    B-Spline-Diff~\cite{qiu2021learning}&0.375$\pm$0.053& 25.338$\pm$4.866& 9.217$\pm$1.960&\textbf{0.0$\pm$0.0}\\
    Fourier-Net \cite{jia2023fourier}&0.417$\pm$0.048 &23.084$\pm$4.201 & 8.000$\pm$1.532 & 2.167$\pm$2.391\\
     \midrule
    LapIRN~\cite{mok2020large}& 0.522$\pm$0.057& 20.728$\pm$4.318&6.274$\pm$1.379&3.203$\pm$0.848 \\
    PRNet++~\cite{kang2022dual}& 0.478$\pm$0.060&23.791$\pm$4.691& 7.524$\pm$1.771& 2.252$\pm$1.192 \\
    ModeT~\cite{modelT} &0.550$\pm$0.054 & 20.351$\pm$4.231&5.784$\pm$1.236& 2.239$\pm$0.906 \\
    Im2Grid~\cite{liu2022coordinate} &0.530$\pm$0.047 &  20.763$\pm$3.786 & 6.028$\pm$1.146 & 2.421$\pm$0.653 \\
    RDN~\cite{hu2022recursive} &0.527$\pm$0.055 & 21.782$\pm$4.722 & 6.461$\pm$1.613 &1.529$\pm$0.989 \\
    \midrule
    Ours~& \textbf{0.588$\pm$0.049} & \textbf{18.253$\pm$3.610}&\textbf{ 5.003$\pm$1.011}&3.930$\pm$1.416\\
            \bottomrule[1pt]
        \end{tabular}
\end{adjustbox}
    \label{tab:ct}
\end{table}

\begin{figure*}[ht]
    \centering
    \includegraphics[width=1\linewidth]{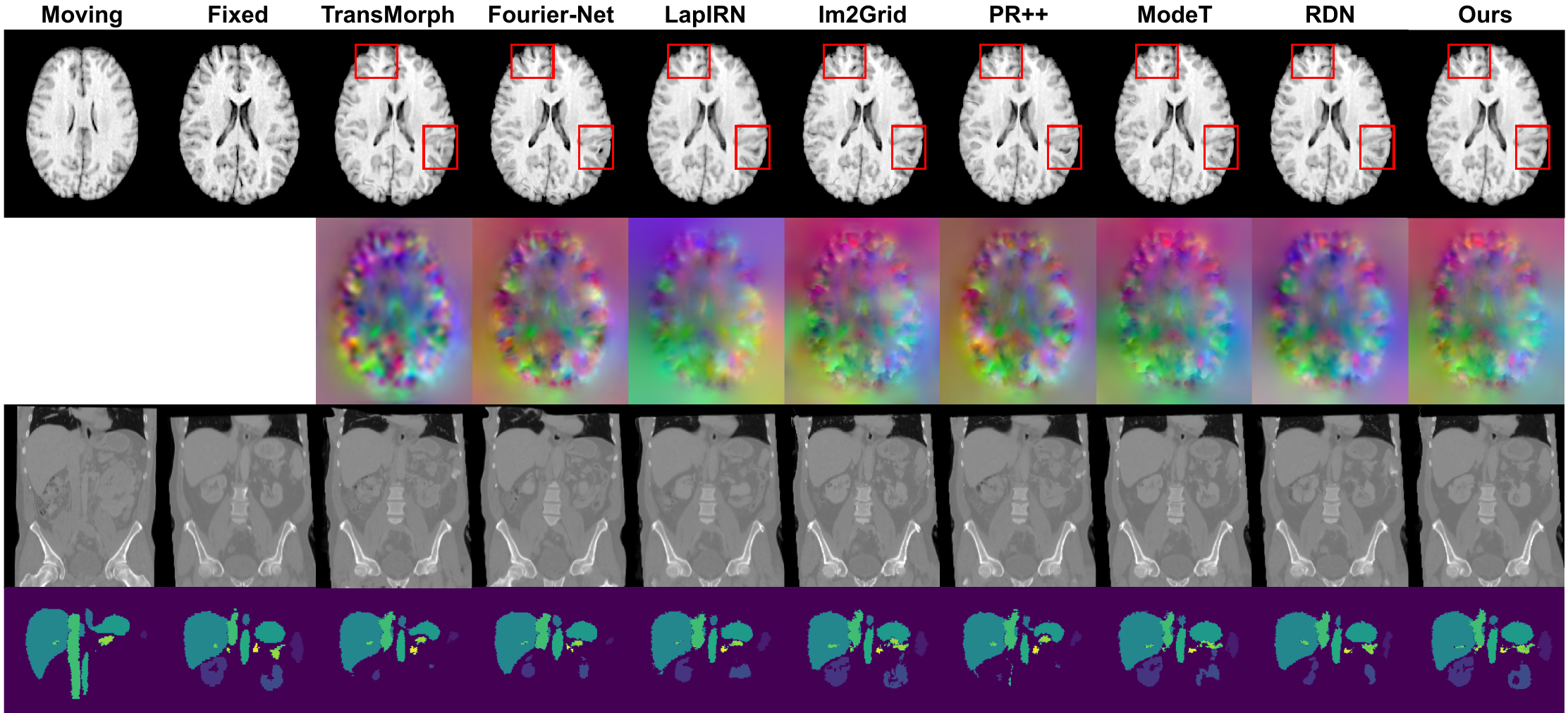}
    \caption{Visual comparisons on the Brain LPBA dataset (top two rows) and the Abdomen CT dataset (bottom two rows). Columns 3–10 display warped moving images (row 1 and 3), displacement fields as RGB images (row 2) and warped segmentation masks (row 4).}
    \label{fig:vis_res1}
\end{figure*}

The visualization of the wrapped moving images and labels, as shown in Fig. \ref{fig:vis_res1}, illustrates that most parts of the kidneys are registered with our method. However, TransMorph, PR++, and RDN failed to register the kidneys, leaving a large portion missing on the warped mask. 
\section{Ablation Study}
\label{sec:ablation}
 \begin{table*}[ht]
\centering
    \caption{Ablation studies on the use of SACB at different scale levels in SACB-Net and varying numbers of spatial clusters ($N$), along with the corresponding average training time (in seconds), including data loading time, and GPU training memory usage (in MiB). The SACB configurations are based on the best results shown in Table \ref{tab:abli_t2}. }
   \begin{adjustbox}{width=1\textwidth}
\begin{small}
        \begin{tabular}{ccccccccccccccccc}
        \toprule[1pt]
       & & \multicolumn{2}{c}{\textbf{With SACB}} &  &\multicolumn{4}{c}{\textbf{IXI}}  &\multicolumn{4}{c}{\textbf{LPBA}}&\multicolumn{4}{c}{\textbf{Abdomen CT}}\\
      \cmidrule(lr){2-5}  \cmidrule(lr){6-9}  \cmidrule(lr){10-13} \cmidrule(lr){14-17} 
         $N$ & Scale$_5$ &Scale$_4$ & Scale$_3$ &Scale$_2$ & Dice $\uparrow$ & $|J|_{ < 0}\%$ $\downarrow$ & Memory & Training & Dice $\uparrow$  & $|J|_{ < 0}\%$ $\downarrow$ & Memory & Training & Dice $\uparrow$  & $|J|_{ < 0}\%$ $\downarrow$ & Memory & Training \\
        \midrule
$\bm{-}$ &\ding{55} &\ding{55} &\ding{55}&\ding{55}&0.7643&0.059 &15838&0.762& 0.7141&0.020&11574&0.545&0.5374&2.907&15838&0.764\\
     \midrule
 5 &\ding{51} &\ding{55} &\ding{55} &\ding{55} & 0.7668 & 0.072&15880 &0.799&0.7217 &0.087&11664&0.579&0.5444  &3.243&15880&0.797\\
 5 &\ding{51} &\ding{51} &\ding{55} &\ding{55} & 0.7679 & 0.076&16104 &0.879&0.7241 &0.104&11770 & 0.631&0.5498  &2.619&16104&0.864\\
 5 &\ding{51} &\ding{51} &\ding{51} &\ding{55} & 0.7671 & 0.060&16960&0.945&0.7266 &0.076&12418&0.714&0.5685  &2.781&16960&0.950\\
 5 &\ding{51} &\ding{51} &\ding{51} &\ding{51} & 0.7683 & 0.086& 21638 & 1.255&0.7294 &0.017&15748&0.918&0.5849  &3.548&21638&1.234\\
    \midrule
 7 &\ding{51} &\ding{51} &\ding{51} &\ding{51} & 0.7691 & 0.083&21638&1.376&0.7309 &0.018&15778 &1.084&0.5881 &3.930&21662&1.360\\
 9 &\ding{51} &\ding{51} &\ding{51} &\ding{51} & 0.7685 & 0.069&21680 & 1.550&0.7304 &0.016&15792&1.110&0.5875 &4.405&21680&1.562\\\
11 &\ding{51} &\ding{51} &\ding{51} &\ding{51} & 0.7684 & 0.084&21732&1.679&0.7300 &0.015& 15806 &1.193 &0.5875 &4.419&21718&1.728\\
    \bottomrule[1pt]
        \end{tabular}
\end{small}
\end{adjustbox}
    \label{tab:abl1}
\end{table*}

First, we note that $\overline{p}(\cdot)$ in Eq.~\eqref{p} can also be defined on the channel dimension or both the spatial+channel (Mix) dimension. As shown in Table \ref{tab:abli_t2}, using the spatial mean of unfolded patches achieves slightly better performance compared to the channel mean and the Mix setting.


We then determine the optimal number of scales at which SACB should be applied and the ideal number of clusters to achieve accurate registration performance. As shown in Table~\ref{tab:abl1}, the optimal configuration is applying SACB from scale$_5$ to scale$_2$.
Additionally, increasing the $N$ from 5 to 7 boosts the performance while a too-large value slightly decreases performance (e.g. 11).

\begin{table}[h]
\centering
    \caption{Ablations of the SACB configurations. After padding and unfolding on the feature map $\mathbf{F}$, the cluster maps can be calculated based on spatial mean on the local unfolded patch, channel mean and Mix (spatial+channel). In this study, we set $N=5$ and apply SACBs from scale$_5$ to scale$_2$. }
\begin{adjustbox}{width=0.35\textwidth}
    \begin{small}
        \begin{tabular}{ccccc}
        \toprule[1pt]
     \multicolumn{2}{c}{\textbf{SACB}} & &\multicolumn{2}{c}{\textbf{LPBA}}  \\
      \cmidrule(lr){1-3}  \cmidrule(lr){4-5} 
      Spatial & Channel & Mix   & Dice $\uparrow$ & $|J|_{ < 0}\%$ $\downarrow$ \\
    \midrule
    \ding{51} &\ding{55} &\ding{55}  &0.7294& 0.017 \\
    \ding{55} &\ding{51} &\ding{55}&  0.7281&0.020\\
    \ding{55} &\ding{55} &\ding{51}&0.7285 &0.021 \\
    \bottomrule[1pt]
        \end{tabular}
            \end{small}
\end{adjustbox}
    \label{tab:abli_t2}
\end{table}
\section{Conclusion}
We introduced SACB-Net, which incorporates a 3D Spatial Awareness Convolution Block (SACB) that adaptively generates spatial convolution kernels based on the feature clusters, thereby improving the network’s ability to capture spatially varying information. Additionally, our proposed flow estimators combine SACB with a similarity-matching mechanism to estimate displacements in a pyramid coarse-to-fine manner. Results on two brain datasets and one abdomen dataset demonstrate that SACB-Net effectively handles both small local and large deformations. Furthermore, the proposed flow estimator is a plug-in module that can be applied in other networks for better feature representation learning.

{\smallskip \noindent\bf{Acknowledgements.}} The research was conducted using the Baskerville Tier 2 HPC service, which was funded by the EPSRC and UKRI through the World Class Labs scheme (EP/T022221/1) and the Digital Research Infrastructure programme (EP/W032244/1). Baskerville is operated by Advanced Research Computing at the University of Birmingham.

{
    \small
    \bibliographystyle{ieeenat_fullname}
    \bibliography{main}
}





\clearpage
\setcounter{page}{1}
\maketitlesupplementary
\section{The Encoder Architecture}\label{sec:supp1}
\begin{figure}[!ht]
    \centering
    \includegraphics[width=0.95\linewidth]{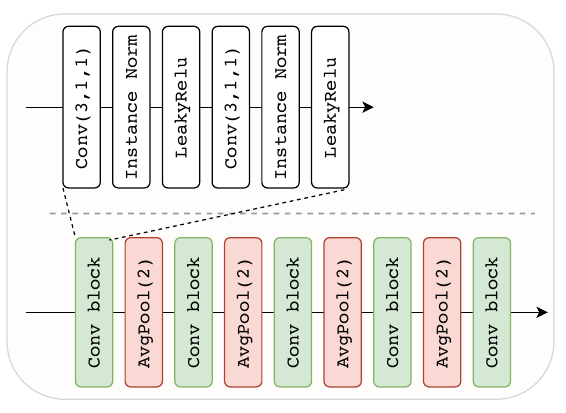}
    \caption{Diagram of the shared encoder architecture, featuring five convolutional blocks to extract multi-scale feature maps and four average pooling layers for downsampling.}
    \label{fig:enc_ar}
\end{figure}

Figure \ref{fig:enc_ar} illustrates the architecture of the shared encoder in SACB-Net. The multi-scale features extracted from each convolutional block are used for pyramid flow estimation.
\section{Normalized Cross-Correlation Loss}
The normalized cross-correlation loss denotes as
\begin{small}
     \begin{align} \nonumber
     &\mathcal{L}_{\text{NCC}}(I_f, I_m \circ \phi) = \\
     &-\sum_{p \in \Omega} \frac{\sum_{p_i} \left( I_f(p_i) - \overline{I_f}(p) \right) \left( I_m \circ \phi(p_i) - \overline{I_m \circ \phi}(p)\right)}{\sqrt{\sum_{p_i} \left( I_f(p_i) - \overline{I_f}(p) \right)^2 \sum_{p_i} \left( I_m \circ \phi(p_i) - \overline{I_m \circ \phi}(p)\right)^2}},
 \end{align} 
 \end{small}where $\overline{I_f}(p)$ and $\overline{I_m \circ \phi}(p)$ denote the local mean intensity values of the images. Here, $p_i$ represents the positions within a local $w^3$ window centered at $p$. During training, we set the window size $w$ to 9.

\section{Evaluation on SSIM metric}
Table \ref{tab:tab_ssim} presents the results of Structural Similarity Index Measure (SSIM) for the comparison methods. However, it has been highlighted by \cite{rohlfing2011image} that a higher degree of image similarity does not always indicate improved registration; anatomical structures are more reliable measures. 
\begin{table}[!ht]
\centering
    \caption{ SSIM$\uparrow$ results.}
    \begin{adjustbox}
    {width=0.35\textwidth}
        \begin{tabular}{lcc}
        \toprule[1pt]
        Method &IXI& LPBA    \\
        \hline
        Affine  &0.680$\pm$0.012 & 0.716$\pm$0.027\\
     \hline
    VM-1~\cite{balakrishnan2019voxelmorph}  &{0.896$\pm$0.012}&{0.940$\pm$0.012} \\
    VM-2~\cite{balakrishnan2019voxelmorph} & {0.900$\pm$0.012}&{0.944$\pm$0.012} \\
    NCA-Morph~\cite{ranem2024nca} &{0.880$\pm$0.016} &{0.922$\pm$0.014} \\
    LKU~\cite{jia2022u}  &{0.860$\pm$0.015}&{0.949$\pm$0.012}\\
     \hline
    B-Spline-Diff~\cite{qiu2021learning}  &{0.858$\pm$0.015}&{0.887$\pm$0.023} \\
    Fourier-Net~\cite{jia2023fourier} &{0.841$\pm$0.016}&{0.908$\pm$0.017}\\
     \hline
    LapIRN~\cite{mok2020large} &{0.898$\pm$0.013}&{0.940$\pm$0.013}  \\
    PRNet++~\cite{kang2022dual}  &{0.929$\pm$0.011}&{0.959$\pm$0.012} \\
    ModeT~\cite{modelT} & {0.922$\pm$0.012}&{0.960$\pm$0.010}\\
    Im2Grid~\cite{liu2022coordinate}  &{0.890$\pm$0.015} &{0.953$\pm$0.011} \\
    RDN~\cite{hu2022recursive} &{0.906$\pm$0.011} &{0.950$\pm$0.011}  \\
     \hline
    Ours&{0.915$\pm$0.012}&{0.965$\pm$0.011}  \\
            \bottomrule[1pt]
        \end{tabular}
\end{adjustbox}
    \label{tab:tab_ssim}
\end{table}
\section{Discussion on LPBA dataset}
\begin{figure}[!ht]
    \centering
\includegraphics[width=\linewidth]{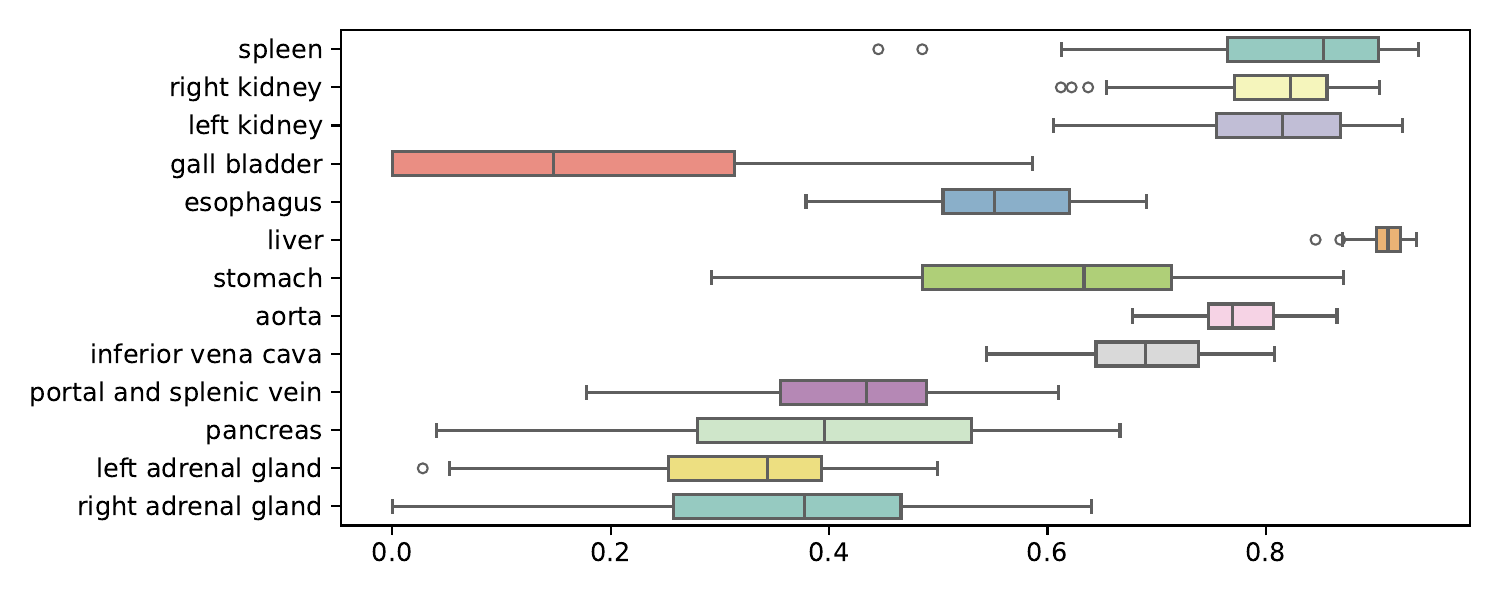}
    \caption{The boxplot of Dice scores for 13 labeled organs.}
    \label{fig:boxplot}
\end{figure}

Figure \ref{fig:boxplot} presents the boxplot of Dice scores for different organs. It is clear that organ size significantly influences registration performance, with the gallbladder being the smallest and most challenging to register, while the liver is the easiest.

\begin{figure}[!ht]
    \centering
\includegraphics[width=\linewidth]{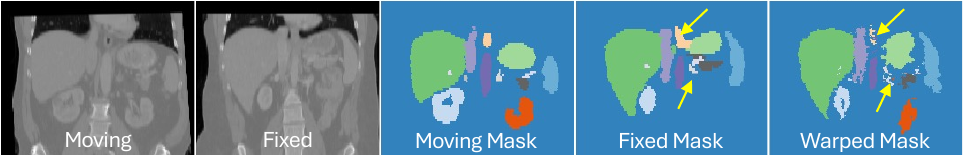}
    \caption{Illustration of a failure case registration with Dice ($<$0.5).}
    \label{fig:failure}
\end{figure}

We selected the case with the lowest average Dice ($<$0.5) as a failure case and presented in Figure \ref{fig:failure}. As shown, small organs are prone to mismatches, significantly impacting registration accuracy.
        
    
    
    
\section{Additional Results}
\begin{figure*}[!ht]
    \centering
    \includegraphics[width=0.9\linewidth]{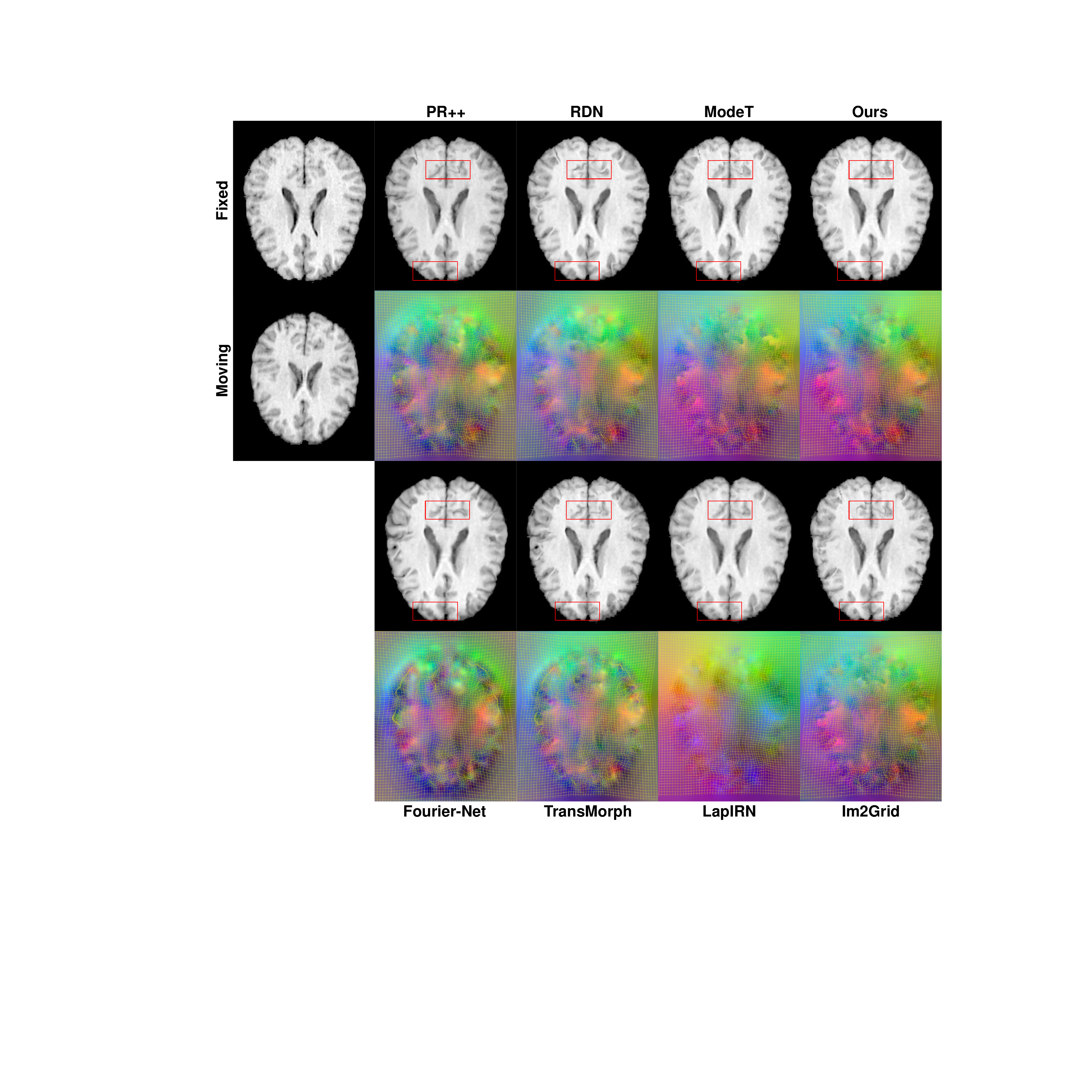}
    \caption{Visual comparisons on LPBA dataset. Columns 2-5: warped moving images (top), displacement fields as RGB images (bottom).}.
    \label{fig:sup_1}
\end{figure*}

\begin{figure*}[!ht]
    \centering
    \includegraphics[width=0.99\linewidth]{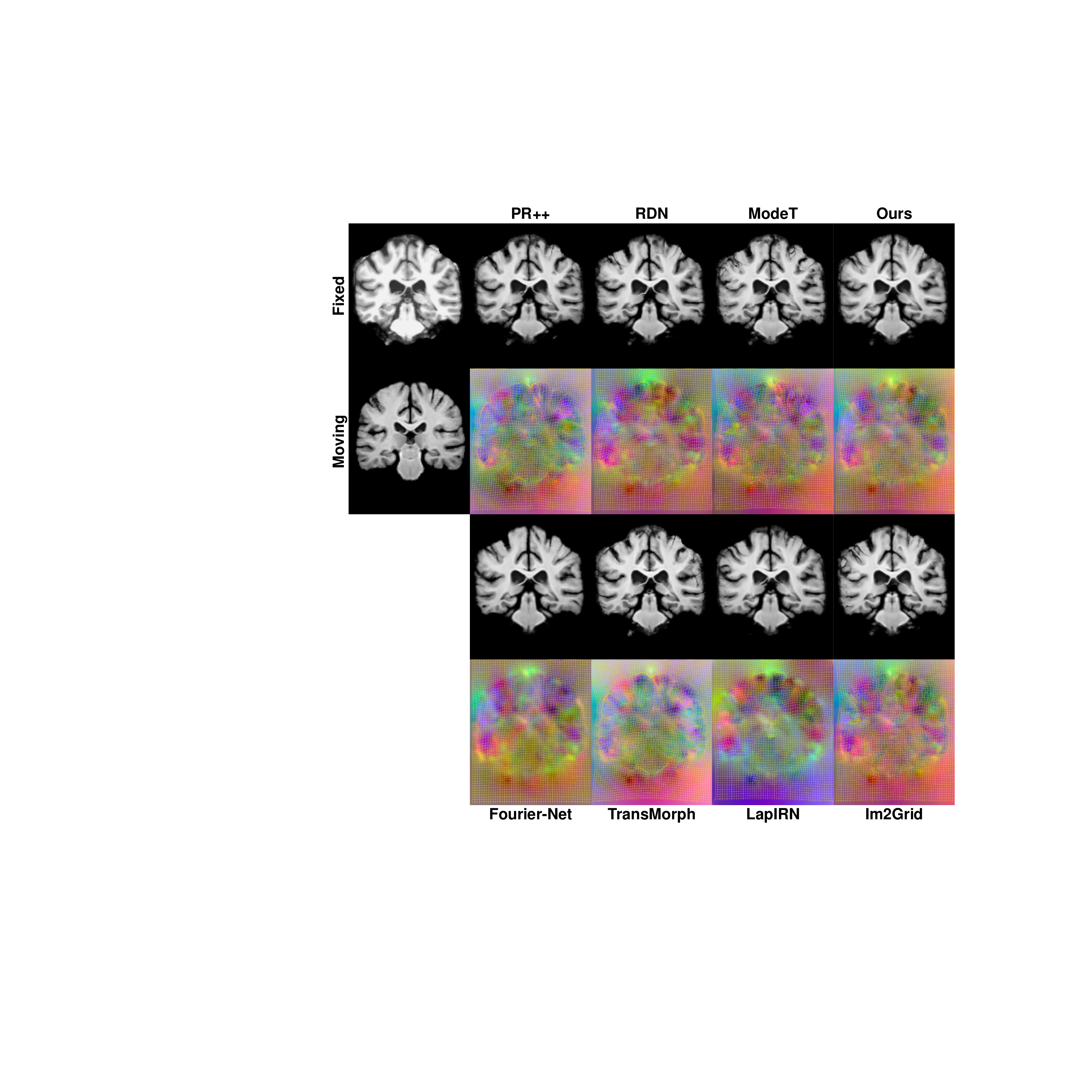}
    \caption{Visual comparisons on IXI dataset. Columns 2-5: warped moving images (top), displacement fields as RGB images (bottom).}
    \label{fig:sup_2}
\end{figure*}

\begin{figure*}[!ht]
    \centering
    \includegraphics[width=0.99\linewidth]{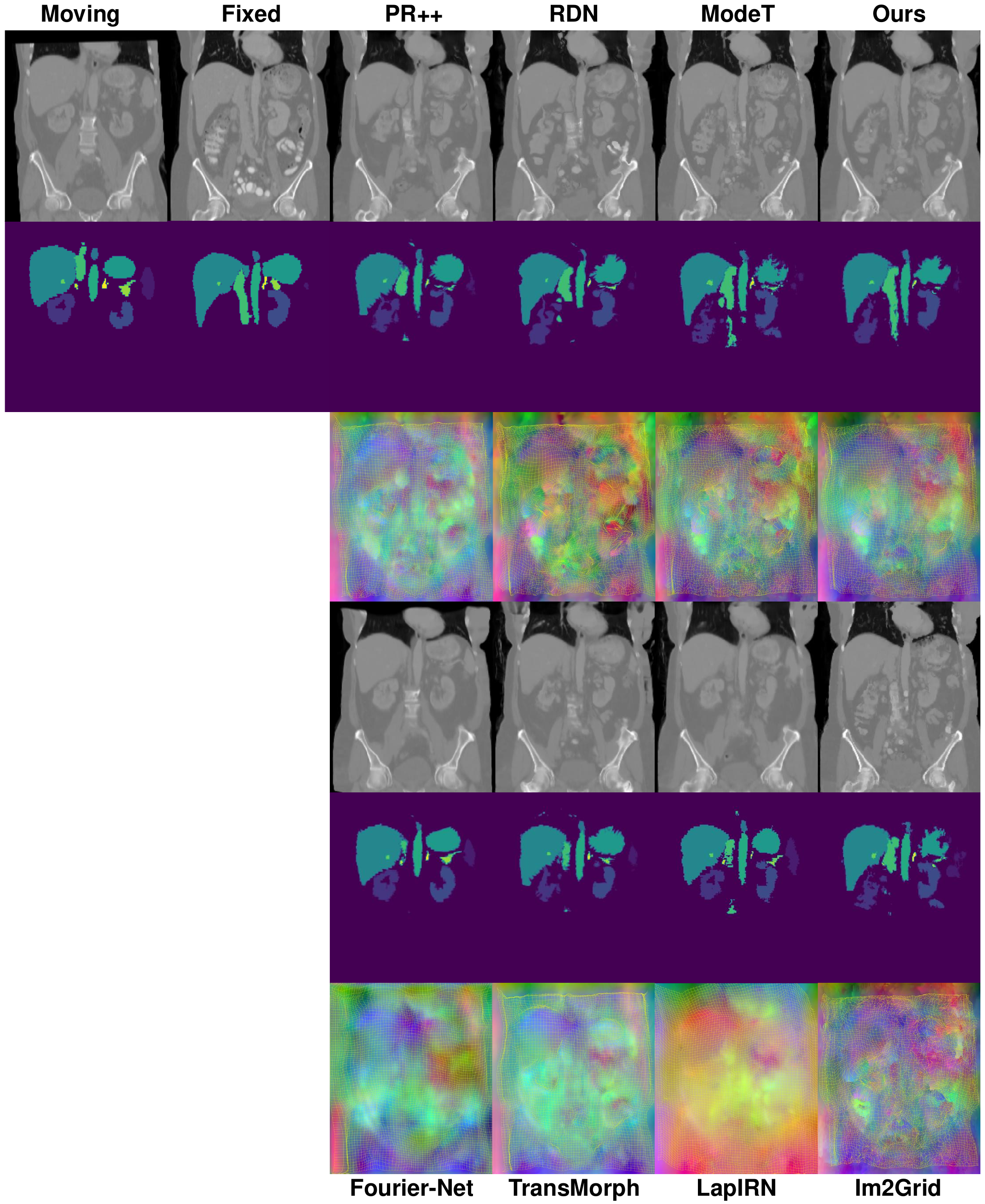}
    \caption{Visual comparisons on Abdomen CT dataset. Columns 3-6: warped moving images (top), warped moving segmentation masks (middle) and displacement fields as RGB (bottom).}
    \label{fig:sup_3}
\end{figure*}

Figures \ref{fig:sup_1}, \ref{fig:sup_2}, and \ref{fig:sup_3} present additional visualization results for the LPBA, IXI and Abdomen CT datasets, respectively, as shown on the following pages.
\newpage
\newpage
\newpage



\end{document}